# Analysis Tool for UNL-Based Knowledge Representation

## Shamim Ripon[1], AoyanBarua, and Mohammad Salah Uddin

Department of Computer Science and Engineering

East West University, Dhaka, Bangladesh

Email: [1]dshr@ewubd.edu

## Abstract

The fundamental issue in knowledge representation is to provide a precise definition of the knowledge that they possess in a manner that is independent of procedural considerations, context free and easy to manipulate, exchange and reason about. Knowledge must be accessible to everyone regardless of their native languages. Universal Networking Language (UNL) is a declarative formal language and a generalized form of human language in a machine independent digital platform for defining, recapitulating, amending, storing and dissipating knowledge among people of different affiliations. UNL extracts semantic data from a native language for Interlingua machine translation. This paper presents the development of a graphical tool that incorporates UNL to provide a visual mean to represent the semantic data available in a native text. UNL represents the semanticsof a sentence as a conceptual hyper-graph. We translate this information into XML format and create a graph from XML, representing the actual concepts available in the native language.

## 1. Introduction

Knowledge representation is one of the central and in some way most familiar concept in Artificial Intelligence. Knowledge representation is fundamentally a surrogate, a substitute for the thing itself, used to enable an entity to determine consequences by thinking rather than acting, i.e., by reasoning about the world rather than taking action in it. It is a set of ontological commitments, an answer to the question: "In what terms should I think about the world?" It is a medium of human expression, i.e., a language in which we say things about the world.The simplest way to represent knowledge of an object is the attribute-value representation. An object is characterized by its attributes, each of which having a fixed range of value. A concept then is a category of objects specified by a logical combination of attribute values(Bahrami&Kaviani,2008).

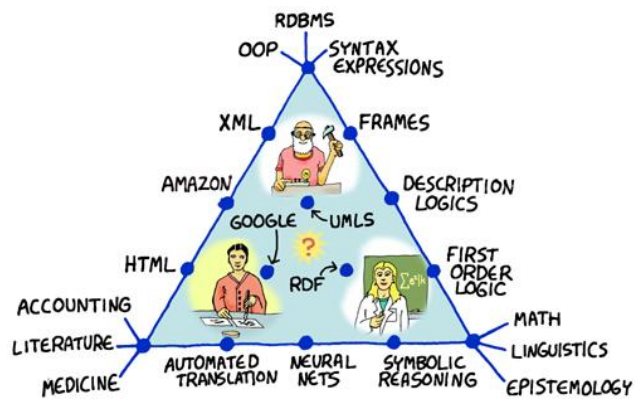

Figure 1: Knowledge representation (http://lisperati.com/tellstuff/rdbms.html)

Fig. 1 illustrates as a triangle the available means to represent knowledge.Using this triangle, we can display all the major ideas in knowledge representation as a function of their pragmatism, scientific rigor, and accessibility to domain experts- The oldest ideas lie at the corners, while more recent ideas appear along the sides. Only in more recent times is software slowly beginning to fill in the areas in the center of this triangle.

XML (Extensible Markup Language) is a universal data interchange format, a web-based storage of concepts and knowledge.XML could serve as a germ of a standardized world-wide knowledge database platform. An XML document consists of elements specified by attribute-value pairs and nested in a hierarchical tree structure. Hence, the idea to apply XML to knowledge representation via the attribute-value representation is straightforward. The logical notions of rules and concepts are naturally mapped into the XML framework (Vries, 2004).The application of XML to concept description has a lot of advantages. Since XML is appropriate for platform-independent and world-wide use XML nowadays has become a widely accepted standard data interchange technology so that its general usability is guaranteed for a long time.

The World Wide Web currently contains a lot of data, various structured data (on-line databases, structured documents) and simple metadata but very little knowledge (Harmelen,&Aifb,1999;Lassila, 1998). Recent fashion in knowledge representation languages is to use XML as the low-level syntax. This tends to make the output of these knowledge representation languages easy for machines to parse, at the expense of human readability(Marinov, 2004). The strength of XML lies in its simplicity to represent data and knowledge. It is used to develop modeling languages that are tailored to specific knowledge, and specific data structures and hierarchies.

A semantic network is often used as a form of knowledge representation. It is a directed graph consisting of vertices representing concepts, and edges representing semantic relation between concepts. An example of semantic network is WorldNet, a lexical database of English. Compared to other more formal networks WorldNet involves fairly loose semantic associations. The main disadvantage is that the semantics is not clear. It is not clear what the representation means, and so the system does not know how to reason or what to infer(Davis,Shrobe,&Szolovits,1993). The disadvantage of semantic networks can be overcome by using XML for Knowledge representation. At first we recognize and collect the objects and instances of objects of the world and draw these items by semantic networks. In the next phase, for better realization and better definition of objects in the world, we use of XML language.Recognition of items and nested relation its essential problem in semantic networks which can be easily handled in XML.

This paper develops a tool to represent knowledge in any native language. Several attempts have been made to represent knowledge in a native language especially in the area of Machine Translation, where the emphasis is on keeping the intended meaning in the native language. To achieve this goal UNL(Uchida& Zhu,1998) provides an approach to represent knowledge based on the relations among the universal words in a sentence which can be represented as a hypergraph. XML on the other hand has been widely used in recent years to express knowledge.

Several tools have been developed for the same. Fig. 2 illustrates the steps that we plan to achieve our goal where our emphasis is in the shaded area of the figure. The native language is first converted into UNL by using an existing Enconverter (Enconverter,2000). A UNL parser is developed to parse the semantic entities in UNL and then XML document isgenerated. A XML generator is developed for this purpose. The generated XML is then parsed to create a graph from the XML document. The graph represents the relationship between the parsed entities in the UNL which actually represents the knowledge in the original native language.

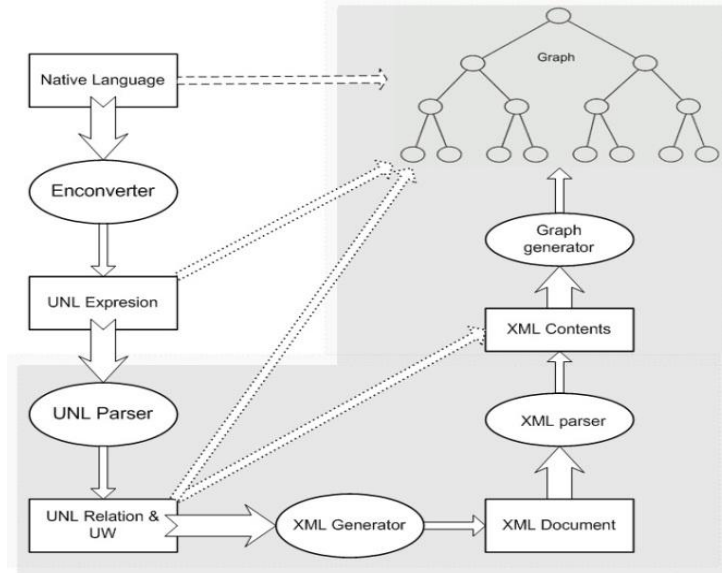

Figure 2 Step-wise project objectives

In the rest of the paper we first give a brief description of UNL along with an example in Section 2. How a UNL expression is parsed and converted into XML document by using the parsed information is illustrated in the following section. The generation of graphical representation from the generated XML document consisting of the parsed information of the original UNL is shown in Section 4. We conclude our paper and outline the strengths and shortcoming of the tool and draw a future plan in Section 5.

## 2. UNL

The UNL is defined as a digital meta language for describing, summarizing, refining, storing and disseminating information in a machine independent and human language neutral form. It represents information sentence by sentence. Each sentence is represented as a hypergraph, where nodes represent concepts and arcs represent relations between concepts. This hypergraph is also represented as a set of directed binary relations between the pair of concepts present in the sentence. Concepts are represented as character-strings called Universal Words (UWs). Knowledge within a UNL document is expressed in three dimensions:

*Universal Words (UWs):*Word knowledge is expressed by Universal Words which are language independent. UWs constitute the UNL vocabulary and the syntactic and semantic units that are combined according to the UNL laws to form UNL expressions. They are tagged using restrictions describing the sense of the word in the current context. For example, `drink(icl>liquor)` denotes the noun sense of `drink` restricting the sense to a type of `liquor`. Here `icl` stands for inclusion and form an 'is-a' relation like in semantic nets(Parikh,Khot,Dave & Bhattacharyya,2004).

*Relation Labels:* Conceptual knowledge is captured by the relationship between Universal Words (UWs) through a set of UNL relations. For example, *"Human affect the environment"* is described in the UNL expression as,

{unl}
agt(affect(icl>do).@present.@entry:01,
human(icl>animal).@pl)
obj(affect(icl>do).@present.@entry:01,
environment(icl>abstract thing).@pl)
{/unl}

where, `agt` means the agent and `obj` means object. The terms `affect(icl>do)`, `human(icl>animal)` and `environment(icl>abstract thing)` are the UWs denoting concepts.

*Attribute Labels:* Speaker's view, aspect, time of event, etc. are captured by UNL attributes. For instance, in the above example, the attribute `@entry` denotes the main predicate of the sentence, `@present` denotes the present tense, `@pl` is for the plural number and `:01` represents the scope ID.The UNL expressions of the sentence *"I went to Malaysia from Bangladesh by aeroplane to attend a conference"* are shown as follows,

```
{unl}
agt(go(icl>move>do,plt>place,plf>place,
agt>thing).@entry.@past,i(icl>person))
plt(go(icl>move>do,plt>place,plf>place,
agt>thing).@entry.@past,
malaysia(iof>asian_country>thing))
plf(go(icl>move>do,plt>place,plf>place,
agt>thing).@entry.@past,
bangladesh(iof>asian_country>thing))
met(go(icl>move>do,plt>place,plf>place,
agt>thing).@entry.@past,
aeroplane(icl>heavier-than-air_craft>thing,
equ>airplane))
obj:01(attend(icl>go_to>do,agt>person,
obj>place).@entry,
conference(icl>meeting>thing).@indef)
pur(go(icl>move>do,plt>place,plf>place,
agt>thing).@entry.@past,:01)
{/unl}
```

## 3. UNL to XML

A number of issues are to be considered while converting UNL expression into corresponding XML document. As one of our objectives is to generate graph representing knowledge in a native language from XML document, structural considerations have to be made while generating XML from UNL. A UNL-XML parser is developed to parse the required entities in UNL. Fig. 3 illustrates the steps to parse the UNL document and to generate XML document.

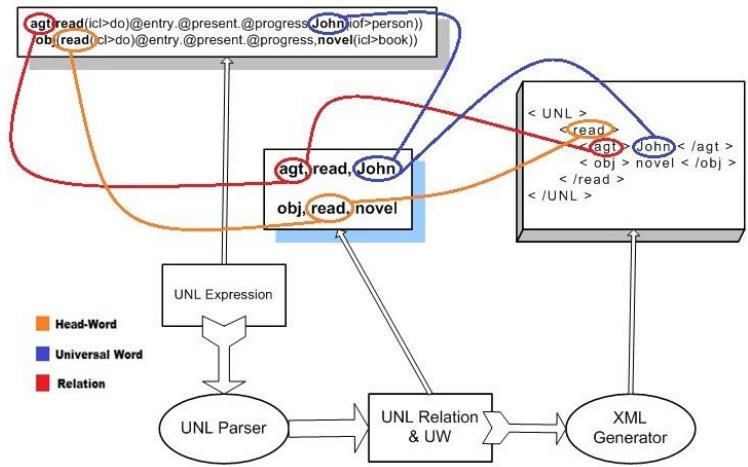

Figure 3 UNL to XML conversion steps

For example, consider the following UNL expression,

```
agt(read(icl>do) @entry.@present.@progress, John(iof>person))
```

UNL parser extracts Relation (agt), Head Word (read) and Universal Word (John) from UNL and these entities will be used to generate the corresponding XML document. Keeping the relationship between the UNL entities the corresponding XML is generated as follows,

```
<UNL> // root
<read> // Element tag
<agt> John </agt> // Content tag & value
</read>
</UNL>
```

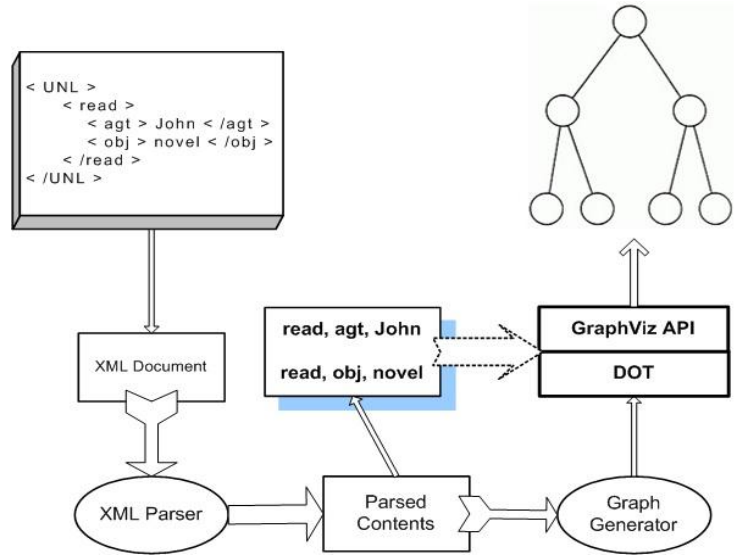

Figure 4 XML-graph conversion steps

## 4. Graph Generation

To generate graph from XML it is required to extract data from XML and then create a graph with those extracted data. DOM parser for java has been used to parse the XML document. After extracting the data from XML, the graph generation tool first organizes the data as parent-child relationship. To draw graphs we use DOT scripting language provided by Graphviz (http://www.graphviz.org/Home.php). The graph generation tool generates a DOT script file from XML document. A graph can then be generated using Graphviz API. The graph generation steps are illustrated in Fig. 4. A screenshot graph generation of developed tool is illustrated in Fig. 5.

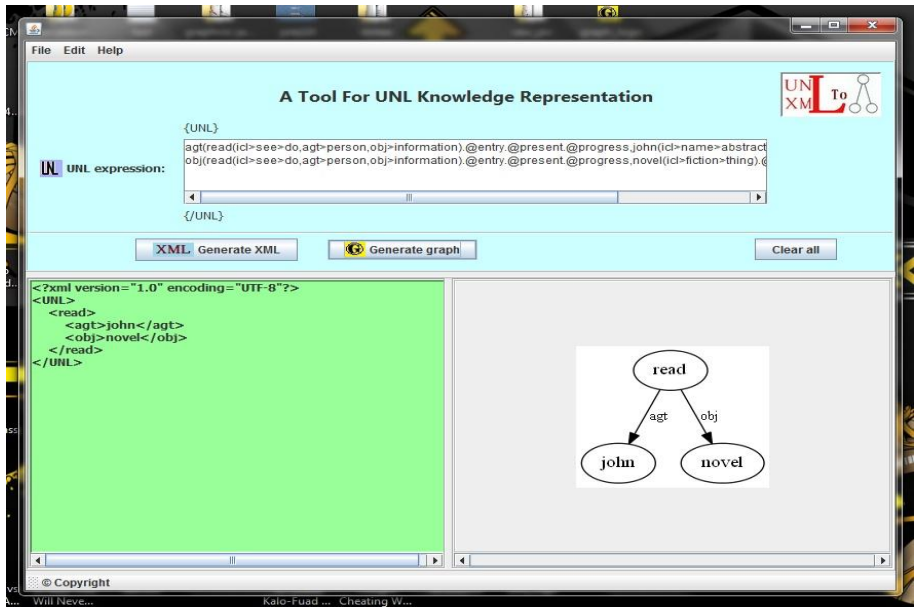

Figure 5 Creating graph from UNL expression

## 5. Conclusion

We have developed a tool to automatically generate XMLdocument and the corresponding graphical representation from UNL expressions. Our primary focus was to represent knowledge in a native language in a graphical form as well as to use a universally accepted platform to represent the knowledge. The tool not only generates graph and XML document but also carries the knowledge entities from UNL to XML preserving the relation (knowledge) of the entities.

UNL has already been widely used as a means of Interlingua machine translation and several works suggested that the UNL can store the intended meaning of the original text. Instead of extracting knowledge form a native language our tool uses UNL to get the required information from a language. The novelty of the present work is to use the benefits provided by UNL to represent knowledge in a widely acceptable format. Whenever knowledgeis represented in XML, it can then be deployed in a variety of applications and several tools have been developed for the same. We have achieved a two-fold result from the tool. Firstly, the tool can be used independently as a UNL parser andsecondly, the XML-graph generator can also act as a separate tool to produce graph from any well-formed XML Document.

Currently our tool takes UNL expression as its input. This UNL is generated by using an Enconverter. It is our plan to interface the Enconverter with our tool so that we can directly take natural input for knowledge representation.We are also interested to incorporate a knowledge management tool that can store the knowledge in a knowledge repository.

Our tool is at its very early stage, where UNL expressions of only simple sentences can be parsed. We are currently improving our tool to parse both complex and compound sentences.For simplicity the tool currently shows the relation between Head-words and Universal-words of UNL. However theattributesdetails are not considered to be included in the presentation. It is our plan to attach the attribute with the nodes of the generated graph that will allow us to check the attribute conveniently from the graph.